\documentclass[10pt,twocolumn,letterpaper]{article}

\usepackage{cvpr}
\usepackage{times}
\usepackage{epsfig}
\usepackage{graphicx}
\usepackage{amsmath}
\usepackage{amssymb}
\usepackage{booktabs}

\usepackage[pagebackref=true,breaklinks=true,letterpaper=true,colorlinks,bookmarks=false]{hyperref}

\cvprfinalcopy 

\def\cvprPaperID{2729} 

\ifcvprfinal\fi
\begin{document}

\title{A Neurobiological Evaluation Metric for Neural Network Model Search}

\author{Nathaniel Blanchard\\
Computer Science and Engineering\\
University of Notre Dame\\
{\tt\small nblancha@nd.edu}
\and
Jeffery Kinnison\\
Computer Science and Engineering\\
University of Notre Dame\\
{\tt\small jkinniso@nd.edu}
\and
Brandon RichardWebster\\
Computer Science and Engineering\\
University of Notre Dame\\
{\tt\small brichar1@nd.edu}
\and
Pouya Bashivan \\
McGovern Institute for Brain Research and \\ Dept. of Brain and Cognitive Sciences 
MIT \\
\tt\small{bashivan@mit.edu} \\
\and
Walter J. Scheirer \\
Dept. of Computer Science and Engineering\\
University of Notre Dame\\
\tt\small{walter.scheirer@nd.edu} \\
}

\maketitle

\begin{abstract}
Neuroscience theory posits that the brain's visual system coarsely identifies broad object categories via neural activation patterns, with similar objects producing similar neural responses. Artificial neural networks also have internal activation behavior in response to stimuli. We hypothesize that networks exhibiting brain-like activation behavior will demonstrate brain-like characteristics, e.g., stronger generalization capabilities. In this paper we introduce a human-model similarity (HMS) metric, which quantifies the similarity of human fMRI and network activation behavior. To calculate HMS, representational dissimilarity matrices (RDMs) are created as abstractions of activation behavior, measured by the correlations of activations to stimulus pairs. HMS is then the correlation between the fMRI RDM and the neural network RDM across all stimulus pairs. We test the metric on unsupervised predictive coding networks, which specifically model visual perception, and assess the metric for statistical significance over a large range of hyperparameters. Our experiments show that networks with increased human-model similarity are correlated with better performance on two computer vision tasks: next frame prediction and object matching accuracy. Further, HMS identifies networks with high performance on both tasks. An unexpected secondary finding is that the metric can be employed during training as an early-stopping mechanism. 
\end{abstract}

\begin{figure}[h]
 \centering
 \includegraphics[width=.475\textwidth]{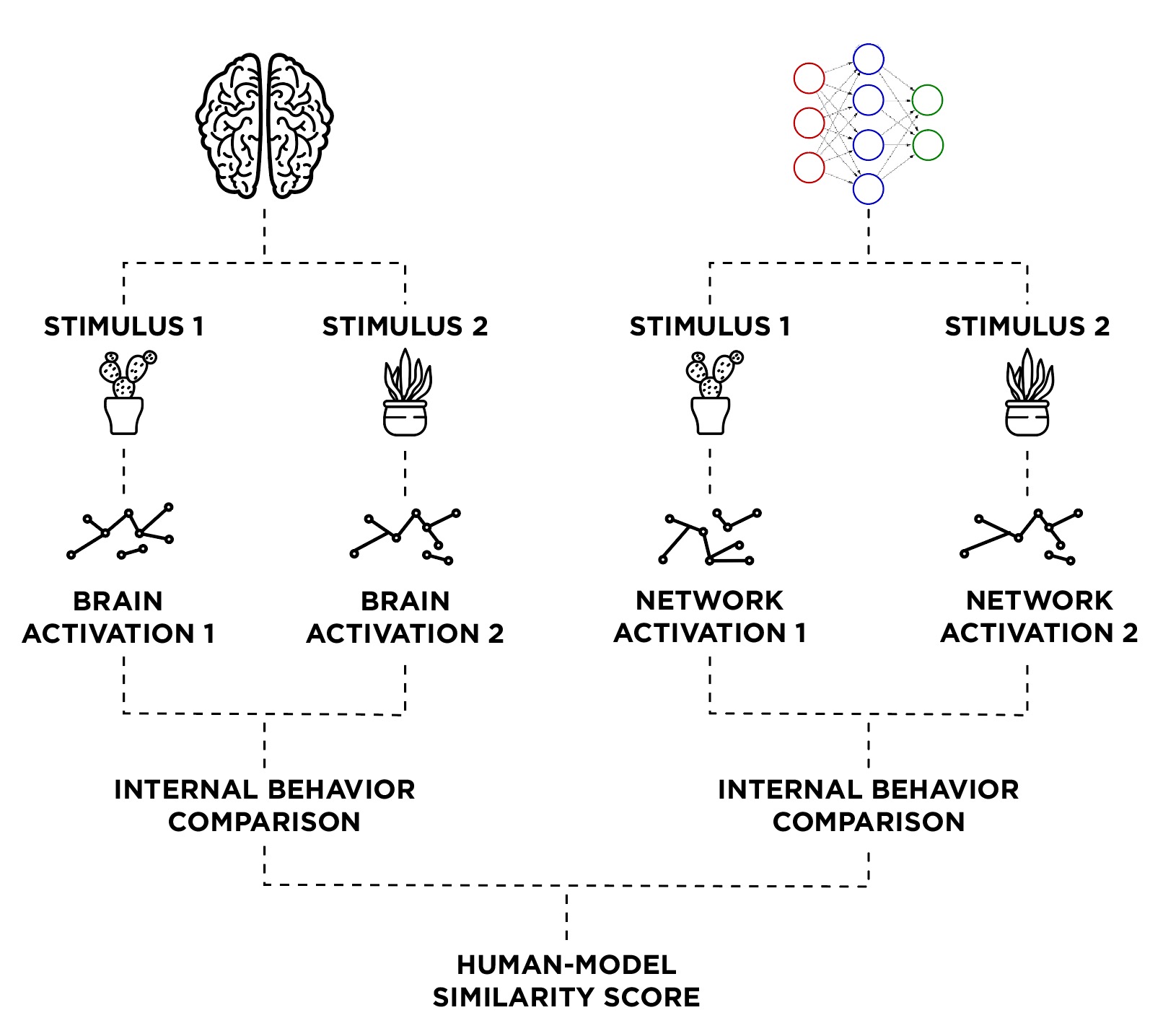}
 \caption{A primary goal of biologically-inspired deep learning work is achieving generalization capabilities that more closely resemble those of biological brains. Along these lines, we propose that model search frameworks for neural network training can be guided by a human-model similarity metric. The metric correlates internal activation behavior of the human brain and neural networks over shared stimuli. In this work, we examine the specific case of fMRI recordings~\cite{kriegeskorte_representational_2008} and predictive coding networks~\cite{lotter_deep_2017}. Internal behavior is measured by the dissimilarity in activations between two stimuli. Human-model similarity is the comparison of internal behavior of a brain and a model on a stimulus set, where higher similarity implies better model generalization.
 }
 \vspace{-3mm}
 \label{teaser_cvpr_2019}
\end{figure}

\section{Introduction}
Researchers originally designed artificial neural networks based on neurobiological structure and function in the hope that such networks would approximate the performance of the biology that inspired them \cite{rosenblatt_perceptron:_1958}. With the advent of modern deep learning techniques, neural networks are finally beginning to realize this original objective across some pattern recognition problems~\cite{lecun2015deep}.
However, we need only consider the learning and processing power of the brain to know that neural network performance is a far stretch from generalized human capabilities~\cite{bonner_computational_2018, geirhos2018generalisation, richardwebster_psyphy:_2016,richardwebster_visual_2018}. This shortcoming has inspired researchers to design new networks which better approximate neurobiological structure, utilizing the architectural elements of machine learning to build networks that embody modern theories of brain organization~\cite{lotter_deep_2017,lotter2018neural,riesenhuber_hierarchical_1999,roper_insect_2017,serre_robust_2007,Yildirim282798}. 
In this paper we look beyond structural similarity and consider behavioral similarity between biological brains and trained networks (\textit{i.e.}, models), as measured by similarity of activation behavior across a set of stimuli. We hypothesize that networks with increased behavioral similarity will exhibit better generalization capabilities across different visual recognition tasks.
 
One neurobiologically-inspired network is the unsupervised predictive coding network~\cite{lotter_deep_2017,rao1999predictive}. Predictive coding networks combine the empirical successes of neural networks with insights from computational neuroscience to train unsupervised models with increased biological fidelity (\textit{i.e.}, the correspondence of an algorithm's representations, transformations, and learning rules with those of their counterparts in the brain). The networks are designed~\cite{lotter_deep_2017} and demonstrated~\cite{lotter2018neural} to embody the theory that in-the-wild biological vision systems are continuously predicting the next input signal~\cite{rao1999predictive}. Additionally, these networks are trained using unsupervised video data, something also done by biological beings~\cite{lecun2015deep}, allowing large-scale unsupervised learning. Finally, the networks have been shown to perform well on at least two different tasks: next frame prediction and object matching~\cite{lotter_deep_2017}. 
 
Predictive coding networks are architecturally designed to emulate neural processing. However, the ability of biological beings to generalize and adapt extends from both structure and internal behavior. Internally, the visual system processes similar objects with similar patterns of cell activations~\cite{coutanche_creatures_2018,garcea2018domain,moss2004anteromedial}. This activation behavior is an observable manifestation of the brain's ability to generalize beyond its experience, such as automatically allowing the classification of unseen instances of object classes (\textit{e.g.}, correctly identifying a car despite never having seen this specific car before). 
We hypothesize that predictive coding networks which mimic the brain's visual behavior will exhibit increased biological fidelity and thus possess strong generalization ability compared to coding networks that do not exhibit this behavior. To test this hypothesis we investigate a new human-model similarity metric (HMS) that evaluates the networks for internal behavioral similarity to fMRI recordings of the human brain (Fig.~\ref{teaser_cvpr_2019}).
 
Systems such as predictive coding networks and biological brains both exhibit internal behavior through their neural activations. Therefore, assessing networks for internal behavior indicative of biological fidelity requires measuring the similarity of activations. To do this, we make use of the recently established technique of representational similarity analysis (RSA) \cite{kriegeskorte_representational_2008,mcclure_representational_2016}. RSA utilizes a set of stimuli to quantify behavioral similarity from activations. For any brain or network, the activation power can be measured in response to a stimulus. The internal behavior can then be defined as the dissimilarity in activations over a set of stimuli. In the case of visual recognition, we expect like-stimuli to have like-activations. We utilize a set of stimuli selected to exhibit a range of both similar and dissimilar objects~\cite{kriegeskorte_matching_2008}. 

\begin{figure*}[h]
 \centering
 \includegraphics[width=.98\textwidth]{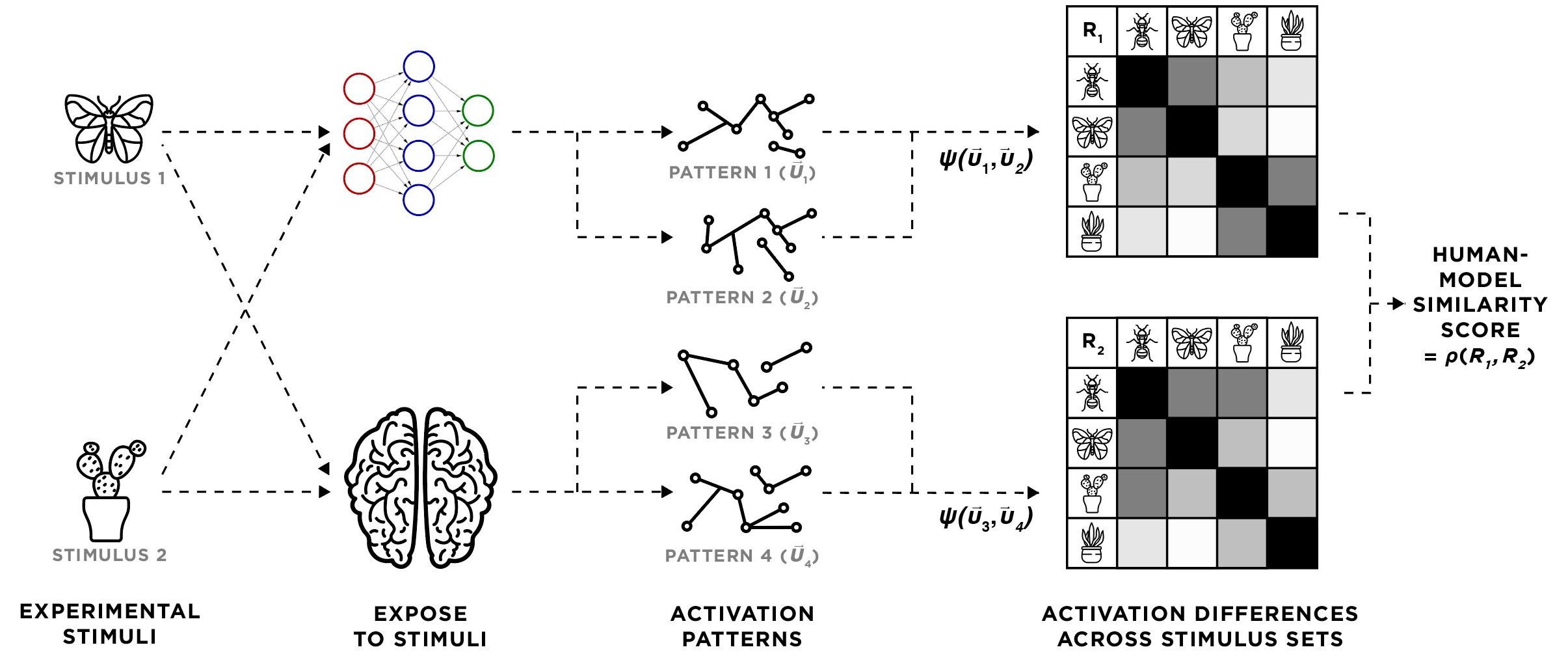}
 \caption{The proposed human-model similarity metric HMS is calculated by comparing the neural activation behavior from two systems: predictive coding networks and fMRI recordings of the human brain. Neural activations are obtained by exposing the systems to stimuli. We abstractly summarize a source based on its internal behavior, generating a similarity score via $\psi$ from activation patterns for each stimulus pair. We then store this internal behavior to the stimuli in an RDM ($R_1$ and $R_2$ above). Finally, the HMS metric $\rho$ is equal to the Spearmen's rank correlation coefficient of the internal behavior of the two sources as measured by the stimulus pairs.}
 \label{method_cvpr_2019}
\end{figure*}

One problem with assessing the similarity of activation behavior of biological beings and neural networks is the absence of a one-to-one mapping between the neurons of the brain and the neurons of the networks. With RSA, complex systems are abstracted into representational dissimilarity matrices (RDMs), composed of the internal behavior of the system, which is the activation dissimilarity over the set of stimuli. The full process to abstract a system into an RDM is illustrated in Fig.~\ref{method_cvpr_2019}. Two input systems can be directly mapped when both are abstracted into RDMs with the same stimuli. Our proposed HMS metric measures human-model similarity as the correlation of a human fMRI RDM and a neural network RDM.
 
We evaluate the HMS metric in a Monte Carlo scenario across a broad range of hyperparameterized networks, data domains, and alternative network metrics. This approach allows us to explore the range of internal behavioral similarity that we can expect to find in predictive coding networks. Additionally, this method allows us to consider how a metric for human-model similarity could be used in the model search process for neural network training. While RSA has been employed to analyze similarities between convolutional neural networks (CNNs) and biological behaviors~\cite{kriegeskorte_relating_2009,mcclure_representational_2016,yamins_hierarchical_2013,yamins_performance-optimized_2014}, a generalized human-model similarity metric as an evaluation of a network's neurbiological fidelity and its use in neural network model search has remained largely untested until now.
Our goal is to present a data-driven study of the HMS metric in order to promote it as a tool for studying generalization in computer vision.

In summary, we make the following contributions: (1) The introduction and evaluation of a new human-model similarity metric, dubbed HMS, to measure network generalizability. (2) The implementation of a metric evaluation framework to assess new machine learning performance metrics. (3) The discovery of HMS as an indicator of a predictive coding network's performance via experiments on the KITTI~\cite{geiger_vision_2013}, VLOG~\cite{fouhey_lifestyle_2017}, and ``Gazoobian Object"~\cite{tenenbaum_how_2011} datasets. (4) The identification of HMS as an early stopping mechanism for training. All code and data associated with this paper will be released following publication.

\section{Related Work}

How to best evaluate machine learning algorithms is an ongoing discussion. Traditional evaluations focus on external performance on a dataset, but there are no guarantees against overfitting or unpredictable network performance on real-world data \cite{torralba2011unbiased}. One alternative evaluation regime is visual psychophysics, which monitors neural network performance while increasingly perturbing stimuli~\cite{leibo2018psychlab,richardwebster_psyphy:_2016, richardwebster_visual_2018}. This evaluation centers on the observation that a network which inconsistently recognizes perturbed stimuli cannot be trustworthy. However, these evaluations are still focused on creating variability within a dataset, offering no guarantee that the network is not simply overfit to it. 
Moving beyond datasets, our proposed evaluation metric HMS quantifies consistency in a network's internal behavior by directly comparing against the internal behavior of one of the most generalizable vision systems in the world: the biological brain~\cite{coutanche_creatures_2018,garcea2018domain,moss2004anteromedial}. 

HMS uses human participant fMRI data as ground-truth internal behavior that leads to good generalization. The comparison between networks and human fMRI data is inspired by Kriegeskorte \etal\cite{kriegeskorte_representational_2008}, who described how network or neural activations could be abstracted into an RDM. An RDM is an abstract representation that can be directly compared against another RDM, as long as both are created from a joint set of stimuli. Fig.~\ref{method_cvpr_2019} shows how internal behavior is calculated and abstracted into RDMs, and how RDMs can be compared. Sec.~\ref{rdm_methods} describes the formal RDM creation process. Kriegeskorte has a long history of utilizing RDMs to study neural behavior~\cite{kriegeskorte_relating_2009,kriegeskorte_pattern-information_2011,kriegeskorte_representational_2008,kriegeskorte_matching_2008,mur_human_2013,nili_toolbox_2014}.

 With respect to the intersection between neuroscience and machine learning, the neuroimaging technique of fMRI has been used as ground-truth for designing features~\cite{Daugman:85}, interpreting neural network features~\cite{kalfas2018representations,long_role_2018}, and studying network performance~\cite{sacramento2018dendritic}. Fong \etal\cite{fong_using_2017} recently found that raw fMRI data could be used to weight support vector machines to improve performance, indicating that coarse-level brain data can potentially help machine learning networks generalize. The success of that study, alongside the public release of human fMRI data in RDM form by Nili \etal\cite{nili_toolbox_2014} further motivated us to use fMRI data as ground-truth in our network evaluation. The specific contributions fMRI data can make in expanding our understanding of neural networks are still to be explored, but to our knowledge this is the first instance of fMRI data being deployed for neural network model search, where the task is to screen different hyperparameter and architecture configurations for models that perform well on a given task. There is significant recent interest in optimization methods, search strategies, and infrastructure for neural network model search~\cite{bashivan_teacher_2018,elsken2018neural,hsu2018monas,liu2018progressive,pham2018efficient}. In this context, our work represents a new capability for such searches.

Extensive research has been performed comparing the neural activity of macaques to CNNs~\cite{hong_explicit_2016,kheradpisheh_deep_2016,yamins_using_2016,yamins_hierarchical_2013,yamins_performance-optimized_2014}. These studies map CNN layers to anatomical visual areas measured with electrode arrays.
Recently, research has shown that these internal representations are not predictive of primate behavior at the image level \cite{rajalingham_large-scale_2018,rosenfeld2018totally}, suggesting CNNs are not mimicking internal behavior well enough.
Given these recent findings, we opted to study more biologically plausible predictive coding networks~\cite{lotter_deep_2017,rao1999predictive}. These networks are unsupervised and are relatively unexplored for many problem domains, but yield state-of-the-art performance for problems such as next frame prediction. We selected the PredNet architecture because of research establishing its emergent properties that are consistent with biological vision~\cite{lotter2018neural}, meaning it is not grounded only in theory. Nonetheless, there are many biologically-inspired neural network architectures~\cite{pinto_high-throughput_2009,riesenhuber_hierarchical_1999,roper_insect_2017,serre_robust_2007,Yildirim282798}, and interest in them continues to grow~\cite{barrett2018analyzing}. All such networks warrant an investigation into internal behavior as well. 

\begin{figure}[h]
 \centering
 \includegraphics[width=.35\textwidth]{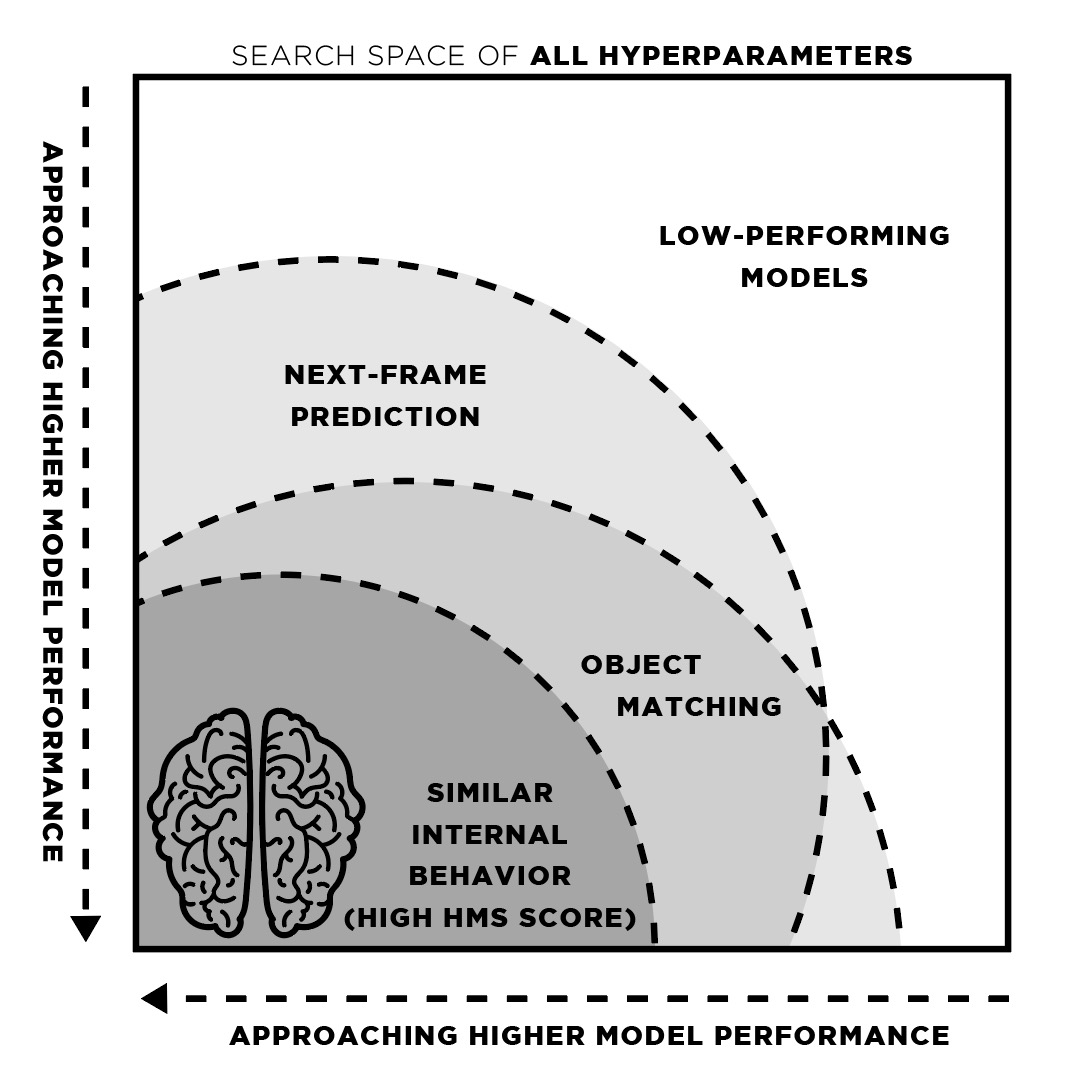}
 \caption{We assess our proposed HMS evaluation metric on randomly hyperparameterized predictive coding networks to study a Monte Carlo-style statistical sample of the space. We evaluate each network with three metrics: HMS, an object matching accuracy metric, and a next frame prediction error metric. We then compare metric performance across the full set of trained networks. We find that networks with higher HMS have high performance on other computer vision metrics, and performance is linked both across and within networks.}
 \label{search_cvpr_2019}
\end{figure}

\section{Methods} 

In this section we introduce the core methodologies surrounding the HMS metric. First, we introduce the biologically-inspired predictive coding network used for the experiments. We then explain the evaluation framework that was used to study HMS, and discuss the computer vision tasks (object matching and next frame prediction) that network performance was evaluated on. Finally, we detail the metric itself (Fig.~\ref{method_cvpr_2019}), explaining: (1) the abstraction of both the fMRI recordings and neural networks into individual RDMs by measuring activations in response to stimuli, and (2) the correlation of the fMRI and the neural network RDMs, which results in the HMS score. 

\subsection{PredNet: A Biologically-Inspired Network} PredNet~\cite{lotter_deep_2017} is a recently introduced unsupervised, biologically inspired, predictive coding network. Its architecture consists of multiple layers (which can vary based on configuration) each incorporating representation neurons (convolutional LSTM units), which output layer-specific predictions at each time step when processing a sequence of data. This output is then compared against a target to calculate an error term, which is propagated laterally and vertically throughout the network.
We follow the training regime laid out by Lotter \etal\cite{lotter_deep_2017}. PredNet is trained without supervision: the network is shown a randomly sampled set of sequential frame sequences and upon viewing each frame, the network attempts to predict the next frame. The network is optimized to reduce the next frame prediction error on the training set.

\subsection{Metric Evaluation Framework} \label{montecarlomethod}
Because we are focused on improving generalizability, we assess the value of HMS as a predictor of other, more standard, performance measures. This involves varying hyperparameters within a network type, obtaining a Monte Carlo-style statistical sample of the search space, and correlating HMS with standard computer vision evaluation metrics across networks in the sample (Fig.~\ref{search_cvpr_2019}).  We analyze the networks by studying the mean, standard deviation, and Spearman's rho (correlation coefficient) of several performance metrics across the set of sampled networks. We ensure significance by reporting Spearman's rho \textit{p} values, which correspond to the likelihood that correlations occur by chance. We also adhere to Cohen's standard recommendation for interpreting effect sizes \cite{cohen1988statistical}, and do not consider small correlations (less than $0.2$) when comparing two different metrics --- even if they reached statistical significance. 
Further, we perform Bonferroni correction, which conservatively adjusts significance to counteract the multiple testing problem where multiple inferences increase the likelihood of erroneous inferences~\cite{dunn1958}. In all of our results, Bonferroni adjusted \textit{p} values are reported. 

In this study, we correlate HMS with mean-squared error (MSE) on the next frame prediction task (the default mode of PredNet), as well as object matching accuracy. 
In the experiments, following the protocol established by Lotter \etal~\cite{lotter_deep_2017}, MSE is computed as the square of the mean pixel-wise difference of the predicted next frame and the actual next frame. Object matching accuracy is evaluated by first extracting the neural activations of the final layer in response to a probe image. Neural activations from the final layer are then extracted across a gallery of $50$ images, one of which is the same object with altered lighting, color, viewing angle, or a combination thereof. Cosine similarity is computed between the probe and the gallery activations, and the gallery image with the highest activation similarity to the probe is the predicted match. 

\subsection{The HMS Metric for Model Search}
Several steps are involved in computing the proposed HMS metric (with the major ones highlighted in Fig.~\ref{method_cvpr_2019}). The RDM creation process described below follows the procedure of the RSA toolbox~\cite{nili_toolbox_2014}. 

\textbf{Stimuli selection for RDM construction.} 
The stimuli were chosen by Kriegeskorte \etal\cite{kriegeskorte_matching_2008} to compare human-primate neural inferior temporal (IT) object representations. The stimuli were selected to provide a hierarchical range of dissimilar and similar objects, such as animate and inanimate objects, not human and human objects, and face and body objects. The full set of stimuli is described in Sec. 1.1 of the Supp. Mat. 

\textbf{Human fMRI collection.} 
The human fMRI data were released as part of the Representational Dissimilarity Toolbox~\cite{nili_toolbox_2014}. All data are in RDM format, meaning we did not directly process the fMRI data, but instead received the set already in usable form. As such, anyone can utilize this data without specific fMRI domain knowledge, which makes the HMS metric broadly applicable to machine learning tasks. Although data from four participants were collected over two sessions, following the methods of Mur \etal~\cite{mur_human_2013}, we averaged the subject RDMs together into a mean human brain RDM, which reduced noise. RDMs were constructed from activations in the bilateral IT region of the brain. 

The full details of the human fMRI data collection can be found in \cite{kriegeskorte_matching_2008}. Nonetheless, for completeness we briefly describe the procedure Kriegeskorte \etal\cite{kriegeskorte_matching_2008} used to collect human fMRI data. Eight RDMs were constructed from fMRI recordings of four subjects over two sessions in response to $92$ stimuli. Recordings were from measurements of $1.95 \times 1.95 \times 2mm^3$ within an occipitotemporal measurement slab ($5cm$ thick). Subjects were presented with a random sequence of the $92$ stimuli. Each stimulus was displayed for $300$ milliseconds, every $3700$ milliseconds, with four seconds between stimuli. Not all voxels were used to construct the RDM. Voxels of interest were selected based on voxel responses to stimuli from an independent dataset. No spatial smoothing or voxel averaging was performed. 

\textbf{PredNet activations to stimuli.} Using the exact same set of $92$ stimuli, we construct an RDM using network activations as features from PredNet's internal representation neurons. Specifically, activations are recorded from the convolutional LSTM units.  Predictive coding networks are time-based networks, and thus we present the stimuli for a fixed five frames and record activations at each time step. We discard the first time step as it corresponds to a ``blank'' prediction. Activation patterns from PredNet for this style of stimuli presentation mimic biological neural responses for perception~\cite{lotter2018neural}. 

\textbf{RDM construction.}\label{rdm_methods} Given a single feature $f$ and a single stimulus $s$, $v = f(s)$, where $v$ is the value of feature $f$ in response to $s$. Likewise, the vector
\begin{align}
	\label{eqn:neurons}
  \vec{v} = \begin{bmatrix}
      v_{1} \\
      v_{2} \\
      \vdots \\
      v_{n}
     \end{bmatrix}^T
     = \begin{bmatrix}
      f_{1}(s) \\
      f_{2}(s) \\
      \vdots \\
      f_{n}(s)
     \end{bmatrix}^T
\end{align}
can represent the feature values of a collection of $n$ features, $f_{1},~f_{2},~...,~f_{n}$, in response to $s$. 
If one expands the representation of $s$ to a set of $m$ stimuli $S = s_{1},~s_{2},...,~s_{m}$, the natural extension of $\vec{v}$ is the set of feature value collections $V = \vec{v}_{1},~\vec{v}_{2},...,~\vec{v}_{m}$, in which $s_{i}\in S$ is paired with $\vec{v}_{i}\in V$ for each $i = 1,2,...,m$. 
The last step prior to constructing an RDM is to define the dissimilarity score between any two $\vec{v}_{i}\in V$ and $\vec{v}_{j}\in V$. We use the symmetric function
\begin{align}
	\label{eqn:ds}
  \psi(\vec{v}_{i},\vec{v}_{j}) &:= 1 - \frac{(\vec{v}_{i}-\bar{v}_{i}) \cdot (\vec{v}_{j}-\bar{v}_{j})}{\| \vec{v}_{i} - \bar{v}_{i} \|_{2}\| \vec{v}_{j} - \bar{v}_{j} \|_{2}}
\end{align}
where $\bar{v}$ is the mean of the features in $\vec{v}$. An RDM $R$ may then be constructed from $S$, $V$, and $\psi$ as:
\begin{align}
	\label{eqn:rdm}
     R 
     = \begin{bmatrix}
      & \psi(\vec{v}_{1},\vec{v}_{2}) & \psi(\vec{v}_{1},\vec{v}_{3}) & \hdots & \psi(\vec{v}_{1},\vec{v}_{m}) \\
      & & \psi(\vec{v}_{2},\vec{v}_{3}) & \hdots & \psi(\vec{v}_{2},\vec{v}_{m}) \\
      & & & \ddots & \vdots \\
      & & & & \psi(\vec{v}_{m-1},\vec{v}_{m}) \\
      & & & &
     \end{bmatrix}
\end{align}

\textbf{Human-model similarity (HMS).} Given any two RDMs $R_{1}$ and $R_{2}$ from the same set of stimuli $S$, one can compute their similarity to determine how similar the activation behavior is in response to $S$. The similarity function
\begin{align}
	\label{eqn:b5}
  HMS = \rho(\hat{R}_{1},\hat{R}_{2}) 
\end{align}
computes a Spearman's rank correlation coefficient represented by $\rho$, where $\hat{R}$ is the flattened RDM. 

Thus HMS is calculated as the correlation between the averaged human fMRI RDM and a constructed PredNet network RDM, obtained from the network activations to the stimuli. The resulting score is defined over the real interval $[-1, 1]$, with $1$ indicating perfect correlation, $-1$ indicating perfect negative correlation, and $0$ indicating the two RDMs are completely uncorrelated. 

\begin{table*}[t]
 \centering
 \begin{tabular}{llr|c}
  \toprule
  \textbf{Evaluation Task} & \textbf{Metric} & \textbf{Mean \hspace{2.55mm} (SD)} & \textbf{Top Ten HMS Mean (SD)} \\
  \midrule
  Next Frame Prediction Error & Pixel MSE  & 0.092 (0.148) & 0.009 (0.003)\\ 
  Object Matching     & Accuracy    & 0.367 (0.134) & 0.459 (0.049)\\
  Human-Model Similarity  & RDM Correlation & 0.106 (0.055) & 0.178 (0.011)\\
  \bottomrule
  \addlinespace[1ex]
 \end{tabular} 
  \caption{A statistical overview of evaluation scores for a sample of $95$ randomly hyperparameterized PredNet networks. These scores indicate the range of scores we expect to obtain from an arbitrary PredNet network. The top ten HMS mean score refers to the average score for each metric for the ten networks with the highest human-model similarity. The top ten average shows that networks with high HMS also achieve high performance on the other tasks. 
 The object matching task was intentionally designed to be difficult --- the network must distinguish fine-grained differences in unseen, fictional ``Gazoobian" objects \cite{tenenbaum_how_2011} where task chance is ($0.02$). Networks are trained using KITTI \cite{geiger_vision_2013} and evaluated on next frame prediction using a held-out set of KITTI data. Pixel MSE is mean squared error of the predicted-to-actual frame at the pixel level. SD is standard deviation.}
\label{training_hyperparameters}
\end{table*}

\section{Experiments}

Our experiments assess how the biological fidelity of predictive coding networks affects two computer vision tasks: next frame prediction and object matching. We identify biological fidelity as more similar internal activation behavior to human fMRI, as measured through RDMs. 

Four datasets are utilized. We evaluate HMS, as described in Sec.~\ref{rdm_methods}, on a dataset of $92$ stimuli with a range of similar and dissimilar objects, from real human faces to animated objects~\cite{kriegeskorte_representational_2008}. Computer vision capabilities are evaluated on two tasks: next frame prediction and object matching accuracy, as described in Sec.~\ref{montecarlomethod}. Next frame prediction is assessed by measuring pixel-level MSE on the KITTI dataset~\cite{geiger_vision_2013}, a video dataset composed of image sequences from a car mounted camera. We also experimented with another video dataset, VLOG~\cite{fouhey_lifestyle_2017}. For object matching, we used a randomly generated ``Gazoobian Objects" dataset (following the procedure described by Tenenbaum \etal\cite{tenenbaum_how_2011}), composed of otherworldly objects guaranteed to be unseen in training. Gazoobian stimuli mirror the stimuli presentation of the HMS stimuli. Even though these objects are well out of domain compared to the natural images used for training, humans are able to generalize to them with ease \cite{tenenbaum_how_2011}, making them an excellent basis from which to study model generalization at inference time. Objects were varied in rotation, lighting, color, or a combination thereof. Example images from all datasets can be found in Sec.~1 of the Supp. Mat.

\subsection{Does HMS Discover Models that Generalize?} \label{training_experiments}

Initially, we evaluated a random Monte Carlo-style sample of hyperparameters in order to test how HMS, next frame prediction, and object matching varied across PredNet networks. In typical model search fashion, we varied six hyperparameters including the number of training epochs, the number of video sequences used for validation after training for an epoch, the number of video sequences used to train within an epoch, the batch size, the learning rate, and the size of the convolutional filters across all layers. The exact space searched is listed in Sec.~2.1 of the Supp. Mat. We trained $95$ $4$-layer PredNets with randomly selected hyperparameters using HyperOpt~\cite{bergstra_making_2013}, a software package for distributed hyperparameter optimization. 

In Table~\ref{training_hyperparameters} we report the mean and standard deviation of the various metrics for the $95$ trained PredNets. Next frame prediction 
was within range of the scores reported by Lotter \etal\cite{lotter_deep_2017}. The accuracy scores highlight the difficulty of the object matching task, which focuses on specific object matching from a $50$ image gallery of stimuli ($\mathrm{chance} = 0.02$). The evaluation scores indicate our parameters were well suited for sampling: performance was above chance but below ceiling. Impressively, the mean HMS was within a standard deviation of the average human-human similarity score of $0.19$ ($\mathrm{SD} = 0.09$). We also confirmed that these results are stable in a cross-dataset context using the VLOG dataset~\cite{fouhey_lifestyle_2017} (these experiments are discussed in Sec.~3 of the Supp. Mat.). 

We next examined how high HMS similarity corresponds with other metrics by looking at the $10$ networks with the highest HMS scores (reported in Table~\ref{training_hyperparameters}). These networks achieve much higher performance over the set of all networks on the two computer vision tasks. We also examined the $10$ networks with the lowest HMS, and note that they have much worse than average performance: mean next frame prediction error was $0.314$ ($\mathrm{SD} = 0.138$), mean object matching accuracy was $0.13$ ($0.15$), and mean HMS was $-0.008$ ($0.027$). This shows HMS is an effective metric for predicting performance. Networks with high HMS perform well, and those with low HMS perform poorly. 

\begin{table*}[t]
 \centering
\begin{tabular}{l|ccc}
  \toprule
  \textbf{Variable}   & \textbf{Accuracy} & \textbf{HMS} & \textbf{Learning Rate} \\
  \midrule
  Next Frame Prediction Error & -0.791\textsuperscript{**} & \bfseries -0.646\textsuperscript{**} & 0.635\textsuperscript{**} \\
  Object Matching Accuracy   & {.} & \bfseries 0.575\textsuperscript{**} & -0.517\textsuperscript{**} \\
  Human-Model Similarity & {.} & {.} & -0.452\textsuperscript{**} \\
  \bottomrule
  \addlinespace[1ex]
  \multicolumn{4}{r}{\textsuperscript{**}$p\,<\,0.001$}
 \end{tabular}
  \caption{Spearman's rho of metrics for 95 trained PredNets with random hyperparameters. The correlations confirm that HMS is predictive of network performance on other metrics. The negative correlation between Next Frame Prediction Error and the two other metrics occurs because next frame prediction is measured by error, which should minimized, while HMS and Accuracy are metrics to be maximized.
  Precautions taken in determining statistical significance are described in Sec.~\ref{montecarlomethod}. Learning rate was correlated with each metric, but was not determined to be a significant contributing factor to HMS as a predictor of network performance after partial correlation analysis.}
\label{Correlating scores}
\vspace{-3mm}
\end{table*}

To be useful, HMS needs to be an effective predictor across all models, not just those that are high performing and low performing. We verified that, across all models, higher HMS is associated with higher performance on the other metrics by computing Spearman's rho across the sampled networks (Table~\ref{Correlating scores}). Further, the $p$ values of these correlations are the probability our findings occur by chance, with $p < 0.001$ indicating a less than 0.001 probability (0.1\%) that our correlations occur by chance (see Sec~\ref{montecarlomethod}. for details of these safeguards). The correlations between the metrics are moderate to strong correlation strengths, with $p < 0.001$. This confirms that HMS is predictive of network performance on computer vision tasks. Additionally, we calculated correlation scores for all hyperparameters to verify that no individual parameter was responsible for these results. We include the learning rate (LR) hyperparameter in Table~\ref{Correlating scores} because it is moderately correlated with the other metrics. 

The correlation with LR indicated a possible risk that LR is strongly influencing the results. We investigated its influence with a partial correlation analysis, which measures the relationship between metrics while controlling for the influence of LR. The correlations between metrics from Table~\ref{Correlating scores} were not statistically significant ($p < 0.001$); however, the sample size was too small for the breadth of LRs tested. We addressed this by repeating the partial correlation experiment on a much larger set of networks ($N = 1811$). For this sample, the correlations between the metrics were statistically significant ($p < 0.001$), with similar correlation strength for the sample. This confirms HMS is significantly correlated with the other metrics regardless of the influence of LR on training. More discussion of this experiment can be found in Sec.~2.2 of the Supp. Mat.

All of the findings discussed above provide evidence that HMS is an effective search metric. HMS was indicative of performance for both computer vision tasks across all models (via correlation) and extremes (top and bottom models). Networks which exhibited more brain-like internal behavior generalized better to other evaluation tasks.

\subsection{Metric Stability During Model Search} \label{stability}

How stable are our evaluation metrics during network training? Do evaluations of network performance vary across identically hyperparameterized models? If HMS fluctuates wildly depending on when one computes it, or due to randomness in training, it may be an unreliable indicator of performance. Through further experimentation, we found that this is not the case, and show that HMS is a more reliable predictor than the other metrics. 

\begin{figure}[b]
 \centering
 \includegraphics[width=.475\textwidth]{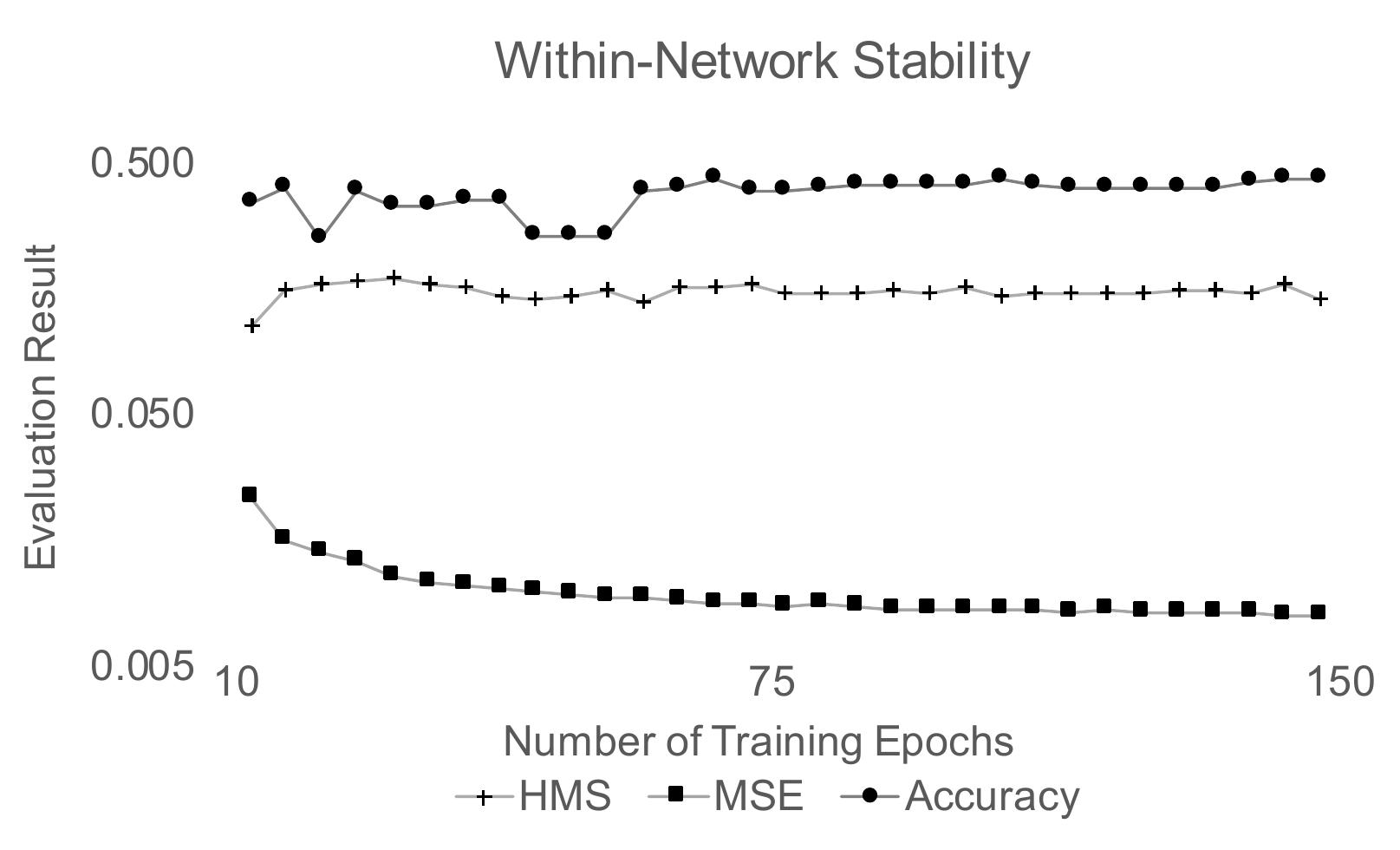}
 \caption{Within-network stability analysis for a representative PredNet model. We find that each metric has its own stereotypical behavior during training. Object matching accuracy is inconsistent early in training, but eventually stabilizes and continues to increase. Next frame prediction error (MSE) either falls consistently (the case shown above) or rises unpredictably, but is heavily dependent on training time. HMS is inconsistent very early in training, but stabilizes more quickly than accuracy, which is unstable for longer, or MSE, which requires a long training time before stabilizing. These findings mean that HMS can be used to identify poor-performing networks for early stopping in network search. 
 }
 \label{within-model}
\end{figure}

\textbf{Within-network stability.}
 We first investigated how the metrics varied during training on a sample of $74$ $4$-layer PredNets trained for $150$ epochs, evaluating performance every $5$ epochs. We focused our analysis on $10$ networks where MSE was below $0.01$ by the $150$th epoch, implying convergence. We found each metric had its own predictable behavior, illustrated by a representative network in Fig.~\ref{within-model}, consistent across hyperparameters. 
 Once HMS was stable ($\mathrm{SD} \le 0.01$) for $25$ epochs it remained so. Object matching accuracy tended to start higher, before dropping, and finally rising again as the training unfolded. Finally, next frame prediction error either continuously decreased, leading to a good network, or increased, leading to a degenerate network. The correlations from Table~\ref{Correlating scores} imply that any of the metrics could be used as a predictor, but the training behavior offers insight into how these metrics would need to be utilized. HMS stabilizes first, after an average of $33$ epochs. Accuracy stabilizes next, after an average of $66.5$ epochs ($\mathrm{SD} = 36$), although some scores did not plateau but continued to increase. In cases where next frame prediction error (MSE) decreased with training, it typically decreased throughout all $150$ epochs, making it the slowest indicator of performance. Note that in the $95$ model sample, MSE was the only metric correlated with the number of epochs ($-0.332, p < 0.001$).  Further details and results can be found in Sec.~5 of the Supp. Mat. 
 
\begin{figure}[t]
 \centering
 \includegraphics[width=.475\textwidth]{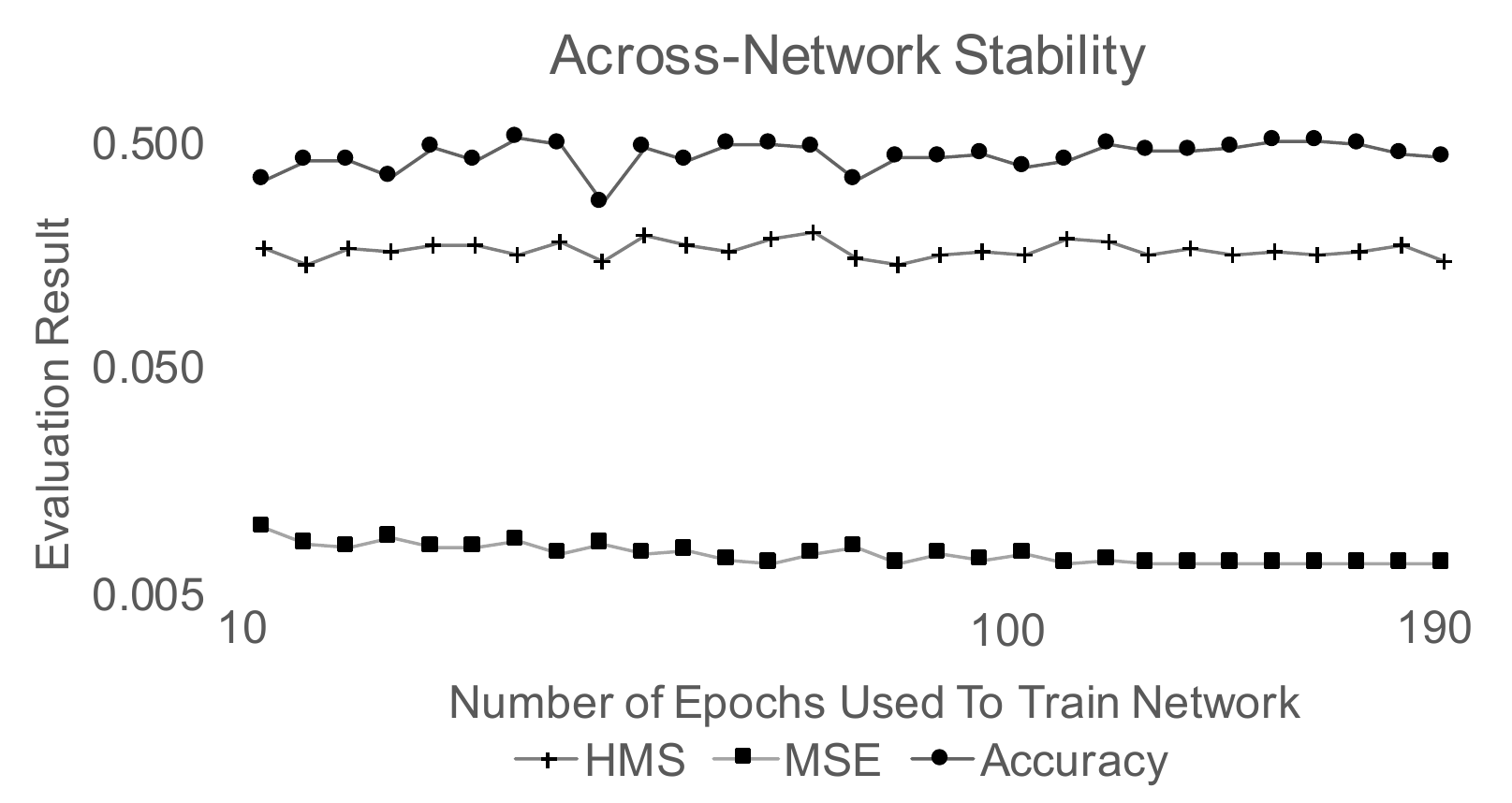}
 \caption{Mean across-network stability for $66$ models trained with a fixed hyperparameter configuration, but for different numbers of epochs. 
  Before $100$ epochs, object matching accuracy is unstable ($\mathrm{SD} = 0.07$) and next frame prediction error (MSE) drops continuously. After $100$ epochs, object matching accuracy is largely stable ($\mathrm{SD} = 0.03$) and MSE largely plateaus, indicating convergence. HMS is the most stable metric ($<100$ epochs $\mathrm{SD} = 0.02$; $>100$ epochs $\mathrm{SD} = 0.01$) and is stable early in training. 
 HMS is thus a stable indication of the potential of a particular combination of hyperparameters across models. All metrics are not strongly impacted by random initialization during training. 
 }
 \label{across-model}
 \vspace{-3mm}
\end{figure}

 \textbf{Across-network Stability.} Next, we evaluated metric stability within a hyperparameter configuration to test how metrics varied with random initialization~\cite{li_convergent_2015}. Ideally, with enough training, the same hyperparameter sets will converge to the same performance, and HMS will be indicative of that performance across those models. We trained $66$ identically hyperparameterized $3$-layer PredNets for a variable number of epochs between $10$ and $500$. We selected a $3$-layer architecture because of its quicker convergence, allowing us to sample more models and see long-term metric behavior. Early in training, behavior mirrored that of the within-network stability experiment. When networks were trained for fewer than $100$ epochs, object matching accuracy was unstable and next frame prediction error continuously decreased. As shown in Fig.~\ref{across-model}, each stabilized around $100$ epochs. However, HMS was consistent across the networks, even in the $10$ epoch setting. Random initialization did not have a strong impact on any metric. Further details and results can be found in Sec.~5 of the Supp. Mat. 
 
\subsection{A Mechanism for Early Stopping}

\begin{figure}[h]
 \centering
 \includegraphics[width=.49\textwidth]{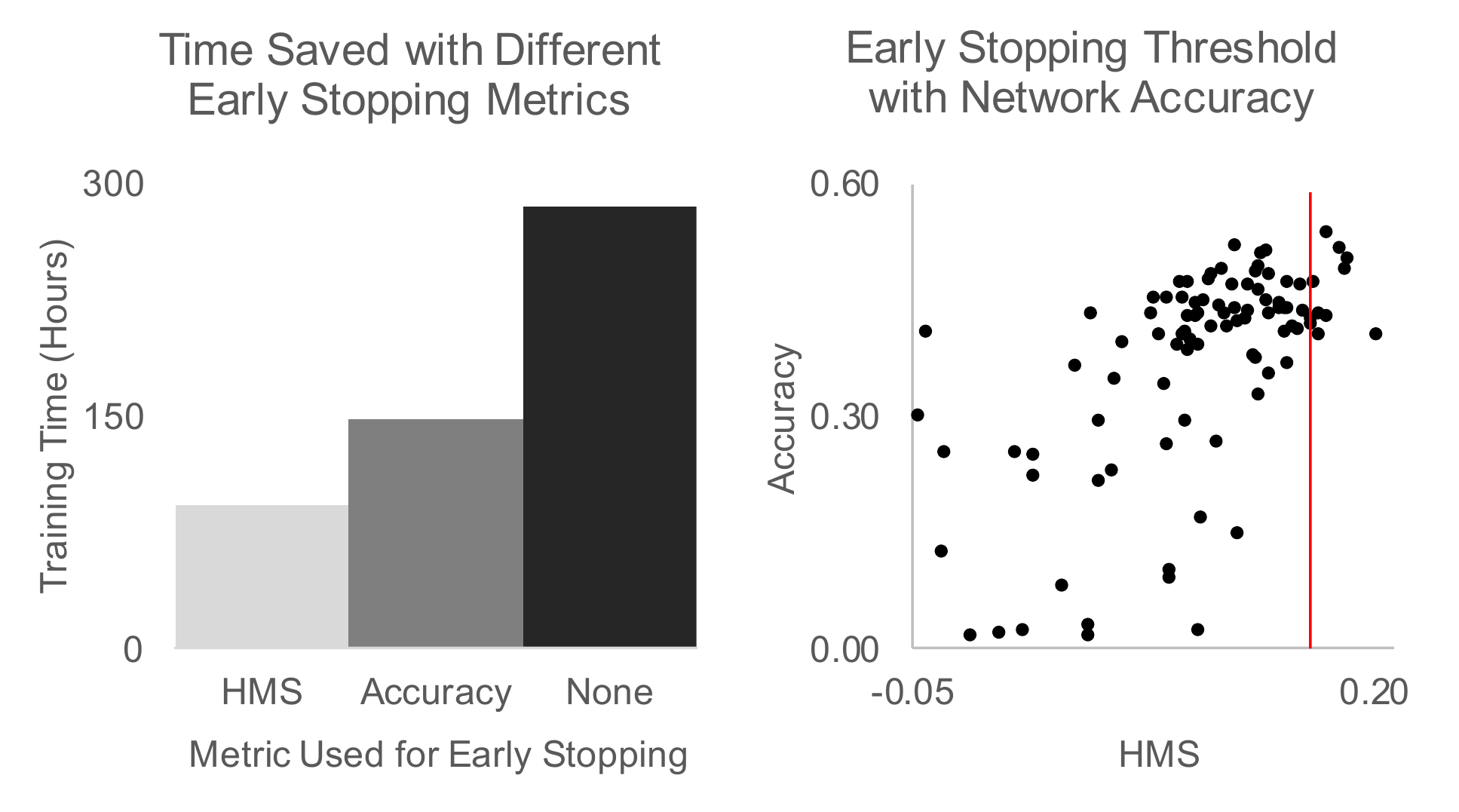}
 \caption{The left-hand plot shows how HMS-driven early stopping for the sample of $95$ PredNets cut down training time by $67$\%, using the criteria for convergence of $\mathrm{SD} \le 0.01$ for 25 epochs. Using the same convergence criteria for accuracy was not as effective. The right-hand scatter plot shows the accuracy of models above and below an early stopping threshold ($0.161$). $82$ models left of the line would be discarded at no cost to final performance. These experiments utilize the findings for metric stability established in Sec.~\ref{stability} to quantify the potential outcome of utilizing early stopping on the model sample.
 }
 \label{early-stopping}
 \vspace{-3mm}
\end{figure}

An outcome of the findings from Sec.~\ref{stability} is that our proposed HMS metric can be employed during network training as a way to discard (\textit{i.e.}, stop training) models that will ultimately perform poorly. To demonstrate this, we conducted a \textit{post hoc} analysis of the $95$ PredNets from Sec.~\ref{training_experiments}. On the left-hand side of Fig.~\ref{early-stopping} we present time saved by early stopping with HMS and accuracy using the convergence criteria of Sec.~\ref{stability}. Overall, early stopping with HMS could have reduced training time by $67$\% at no cost to final performance. We also tested a threshold strategy which considered a network to be stopped during training if its HMS score was below a threshold of $0.161$ (the mean HMS from Table~\ref{training_hyperparameters} plus one standard deviation). Only $13$ of the $95$ models ($13.7$\%) were above this threshold. The right-hand side of Fig.~\ref{early-stopping} depicts the accuracy scores of the models with respect to the side of the HMS threshold they are on. Our analysis shows that even with a high threshold to stop training, and the loss of some models with high performance, most retained models are high performing and were more likely to have high performance on both tasks. Additionally, in this case the highest performance model on both computer vision tasks is retained, but a number of other retained higher performing models have trivial differences in performance, and would be just as useful had the top model been discarded.  Complete details for these experiments and additional results can be found in Sec.~6 of the Supp. Mat. 

\section{Discussion}

One concern that can be raised is the perceived difficulty of obtaining fMRI data. Fortunately, there is a growing open science movement within neuroscience. The fMRI data used in this study is publicly available and can be utilized by anyone~\cite{nili_toolbox_2014}, and it is far from the only data available. Vast public fMRI repositories exist for vision, text, and audio tasks, and researchers do not need to be experts in order to utilize them. A few examples are the Donders repository \cite{fairsharing_team_donders_2018}, OpenNeuro \cite{openneuro}, and Oasis \cite{marcus_open_2010} brains.

Many interesting possibilities exist to extend this research, and one promising direction is recent work showing that RDMs can be incorporated into guided network search by building a ``teacher'' RDM to direct updates~\cite{bashivan_teacher_2018,mcclure_representational_2016}. In these studies, the teacher RDM is constructed from other artificial neural networks, but treating the human fMRI RDM as the teacher is an intriguing possibility, given our findings.

We believe that networks with more biological fidelity in function will be essential to overcome the shortcomings of today's networks in replicating biological vision.
For example, Long and Konkle~\cite{long_role_2018} analyzed CNNs and found that they are driven more by texture than outer contour properties. This finding was echoed by Bonner and Epstein~\cite{bonner_computational_2018} and Laskar \etal\cite{laskar2018correspondence}. Further, Shwartz-Ziv and Tishby~\cite{shwartz2017opening} observed that CNN training largely comprises of compressing input into an efficient representation, with minimal time being used for fitting to training labels. Other aspects of visual processing are likely needed for a better model. 

 The future of artificial intelligence research will need to bridge the gap between network structure and internal behavior, which requires reassessing how we evaluate networks. In the past, unexpected network behavior has blinded-sided researchers, \textit{e.g.}, susceptibility to adversarial images. And it is important to remember that current networks are not consistent with human behavior~\cite{richardwebster_psyphy:_2016,richardwebster_visual_2018}. Evaluations measuring internal behavior should prove useful for avoiding unforeseen issues, and may help us achieve the next generalization breakthrough. 

{\small
\bibliographystyle{ieee}
\bibliography{egbib}
}

\end{document}